\documentclass[10pt,twocolumn,letterpaper]{article}

\usepackage{wacv}
\usepackage{times}
\usepackage{epsfig}
\usepackage{graphicx}
\usepackage{amsmath}
\usepackage{amssymb}

\usepackage[labelformat=simple]{subfig}

\usepackage[pagebackref=true,breaklinks=true,letterpaper=true,colorlinks,bookmarks=false]{hyperref}

\wacvfinalcopy 

\def\wacvPaperID{139} 

\ifwacvfinal\fi
\begin{document}

\title{Local Learning with Deep and Handcrafted Features\\
for Facial Expression Recognition}

\author{Mariana-Iuliana Georgescu$^{1,2}$ \hspace{2cm} {Radu Tudor} Ionescu$^{1,3}$ \hspace{2cm} Marius Popescu$^{1,3}$\\
\\
$^{1}$University of Bucharest, 14 Academiei, Bucharest, Romania\\
$^{2}$Novustech Services, 12B Aleea Ilioara, Bucharest, Romania\\
$^{3}$SecurifAI, 21D Mircea Vod\u{a}, Bucharest, Romania\\
{\tt\small georgescu\_lily@yahoo.com, raducu.ionescu@gmail.com, popescunmarius@gmail.com}
}

\maketitle
\ifwacvfinal\thispagestyle{empty}\fi

\begin{abstract}
We present an approach that combines automatic features learned by convolutional neural networks (CNN) and handcrafted features computed by the bag-of-visual-words (BOVW) model in order to achieve state-of-the-art results in facial expression recognition. To obtain automatic features, we experiment with multiple CNN architectures, pre-trained models and training procedures, e.g. Dense-Sparse-Dense. After fusing the two types of features, we employ a local learning framework to predict the class label for each test image. The local learning framework is based on three steps. First, a k-nearest neighbors model is applied in order to select the nearest training samples for an input test image. Second, a one-versus-all Support Vector Machines (SVM) classifier is trained on the selected training samples. Finally, the SVM classifier is used to predict the class label only for the test image it was trained for. Although we have used local learning in combination with handcrafted features in our previous work, to the best of our knowledge, local learning has never been employed in combination with deep features. The experiments on the 2013 Facial Expression Recognition (FER) Challenge data set, the FER+ data set and the AffectNet data set demonstrate that our approach achieves state-of-the-art results. With a top accuracy of $75.42\%$ on FER 2013, $87.76\%$ on the FER+, $59.58\%$ on AffectNet 8-way classification and $63.31\%$ on AffectNet 7-way classification, we surpass the state-of-the-art methods by more than $1\%$ on all data sets.
\end{abstract}

\maketitle

\section{Introduction}

Automatic facial expression recognition is an active research topic in computer vision, having many applications including human behavior understanding, detection of mental disorders, human-computer interaction, among others. In the past few years, most works~\cite{Barsoum-ICMI-2016,Ding-FG-2017,Giannopoulos-2018,Hasani-CVPRW-2017,Hua-Access-2019,Kim-JMUI-2016,Li-MTA-2017,Li-CVPR-2017,Liu-CVPRW-2017,Li-ICPR-2018,Meng-FG-2017,Mollahosseini-CVPRW-2016,Tang-WREPL-2013,Wen-CC-2017,Yu-ICMI-2015,Zeng-ECCV-2018} have focused on building and training deep neural networks in order to achieve state-of-the-art results. Engineered models based on handcrafted features~\cite{Al-Chanti-VISIGRAPP-2018,Ionescu-WREPL-2013,Shah-PRL-2017,Shao-PRL-2015} have drawn very little attention, since such models usually yield less accurate results compared to deep learning models. In this paper, we show that we can surpass the current state-of-the-art systems by combining automatic features learned by convolutional neural networks (CNN) and handcrafted features computed by the bag-of-visual-words (BOVW) model, especially when we employ local learning in the training phase. In order to obtain automatic features, we experiment with multiple CNN architectures, such as VGG-face~\cite{Parkhi-BMVC-2015}, VGG-f~\cite{Chatfield-BMVC-14} and VGG-13~\cite{Barsoum-ICMI-2016}, some of which are pre-trained on other computer vision tasks such as object class recognition~\cite{Russakovsky2015} or face recognition~\cite{Parkhi-BMVC-2015}. We also fine-tune these CNN models using a novel training procedure known as Dense-Sparse-Dense (DSD)~\cite{Han-ICLR-2017}. To our knowledge, we are the first to successfully apply DSD to train CNN models for facial expression recognition. In order to obtain handcrafted features, we use a standard BOVW model, which is based on a variant of dense Scale-Invariant Feature Transform (SIFT) features~\cite{Lowe-SIFT-2004} extracted at multiple scales, known as Pyramid Histogram of Visual Words (PHOW)~\cite{bosch-phow-2007}. We use automatic and handcrafted features both independently and together. For the independent models, we use either softmax (for the fine-tuned CNN models) or Support Vector Machines (SVM) based on the one-versus-all scheme. For the combined models, the one-versus-all SVM is used both as a global learning method (trained on all training samples) or as a local learning method (trained on a subset of training samples, selected specifically for each test sample using a nearest neighbors scheme). We combine the automatic and handcrafted features by concatenating the corresponding feature vectors, before the learning stage. For the combined models, we explore only global or local SVM alternatives. 
We perform a thorough experimental study on the 2013 Facial Expression Recognition (FER) Challenge data set~\cite{Goodfellow-ICONIP-2013}, the FER+ data set~\cite{Barsoum-ICMI-2016}, and the AffectNet~\cite{Mollahosseini-TAC-2017} data set, comparing our combined deep and handcrafted models with recent and relevant state-of-the-art approaches~\cite{Barsoum-ICMI-2016,Connie-MIWAI-2017,Hua-Access-2019,Ionescu-WREPL-2013,Kim-JMUI-2016,Li-MTA-2017,Li-ICPR-2018,Mollahosseini-TAC-2017,Tang-WREPL-2013,Yu-ICMI-2015,Zeng-ECCV-2018}. We report top results on each and every data set with our combination of automatic and handcrafted features, especially when local SVM is employed in the learning phase. With a top accuracy of $75.42\%$ on the FER 2013 data set, we surpass the state-of-the-art accuracy~\cite{Connie-MIWAI-2017} by $2.02\%$. We also surpass the best method~\cite{Barsoum-ICMI-2016} on the FER+ data set by $2.77\%$, reaching the best accuracy of $87.76\%$. The evaluation on AffectNet is typically conducted using 8-way classification~\cite{Mollahosseini-TAC-2017,Zeng-ECCV-2018} or 7-way classification~\cite{Hua-Access-2019,Li-ICPR-2018} (the class corresponding to \textit{contempt} being removed). We attain the best results in both settings, surpassing Mollahosseini et al.~\cite{Mollahosseini-TAC-2017} by $1.58\%$ in the 8-way classification task, and Hua et al.~\cite{Hua-Access-2019} by $1.20\%$ in the 7-way classification task. We also include ablation results in the paper, which indicate that the proposed model combination yields superior performance compare to each and every component.

Although automatic and handcrafted features have been combined before in the context of facial expression recognition~\cite{Connie-MIWAI-2017,Kaya-IVC-2017}, different from the related art, $(1)$ we include various CNN architectures and a single handcrafted model, and $(2)$ we employ a local learning strategy that leads to superior results. To the best of our knowledge, our previous work based on the BOVW model~\cite{Ionescu-WREPL-2013} is the only one to explore local learning for facial expression recognition. In this paper, we extend our previous work~\cite{Ionescu-WREPL-2013} and propose to combine local learning with automatic features learned by deep CNN models. Compared to the best accuracy reported in~\cite{Ionescu-WREPL-2013} for FER 2013, which is $67.48\%$, we report an improvement of almost $8\%$. In summary, our contributions consist of $(i)$ successfully training CNN models for facial expression recognition using Dense-Sparse-Dense~\cite{Han-ICLR-2017}, $(ii)$ successfully combining automatic and handcrafted features with local learning, $(iii)$ conducting an extensive empirical evaluation with various deep, engineered and combined facial expression recognition models, and $(iv)$ reporting state-of-the-art results on three benchmark data sets.

The rest of the paper is organized as follows. We present recent related art in Section~\ref{sec_RW}. We describe the automatic and handcrafted features, as well as the learning methods, in Section~\ref{sec_M}. We present the experiments on facial expression recognition in Section~\ref{sec_E}. Finally, we draw our conclusions in Section~\ref{sec_C}.

\section{Related Work}
\label{sec_RW}

The early works on facial expression recognition are mostly based on handcrafted features~\cite{Tian-HFR-2011}. After the success of the AlexNet~\cite{Hinton-NIPS-2012} deep neural network in the ImageNet Large Scale Visual Recognition Challenge (ILSVRC)~\cite{Russakovsky2015}, deep learning has been widely adopted in the computer vision community. Perhaps some of the first works to propose deep learning approaches for facial expression recognition were presented at the 2013 Facial Expression Recognition (FER) Challenge~\cite{Goodfellow-ICONIP-2013}. Interestingly, the top scoring system in the 2013 FER Challenge is a deep convolutional neural network~\cite{Tang-WREPL-2013}, while the best handcrafted model ranked only on the fourth place~\cite{Ionescu-WREPL-2013}. With only a few exceptions~\cite{Al-Chanti-VISIGRAPP-2018,Shah-PRL-2017,Shao-PRL-2015}, most of the recent works on facial expression recognition are based on deep learning~\cite{Barsoum-ICMI-2016,Ding-FG-2017,Giannopoulos-2018,Hasani-CVPRW-2017,Hua-Access-2019,Kim-JMUI-2016,Li-MTA-2017,Li-CVPR-2017,Liu-CVPRW-2017,Li-ICPR-2018,Meng-FG-2017,Mollahosseini-CVPRW-2016,Wen-CC-2017,Yu-ICMI-2015,Zeng-ECCV-2018}. Some of these recent works~\cite{Hua-Access-2019,Kim-JMUI-2016,Li-MTA-2017,Wen-CC-2017,Yu-ICMI-2015} proposed to train an ensemble of convolutional neural networks for improved performance, while others~\cite{Connie-MIWAI-2017,Kaya-IVC-2017} combined deep features with handcrafted features such as SIFT~\cite{Lowe-SIFT-2004} or Histograms of Oriented Gradients (HOG)~\cite{Dalal-HOG-2005}. While most works studied facial expression recognition from static images, some works tackled facial expression recognition in video~\cite{Hasani-CVPRW-2017,Kaya-IVC-2017}. Hasani et al.~\cite{Hasani-CVPRW-2017} proposed a network architecture that consists of 3D convolutional layers followed by a Long-Short Term Memory (LSTM) network that together extract the spatial relations within facial images and the temporal relations between different frames in the video. Different from other approaches, Meng et al.~\cite{Meng-FG-2017} and Liu et al.~\cite{Liu-CVPRW-2017} presented identity-aware facial expression recognition models. Meng et al.~\cite{Meng-FG-2017} proposed to jointly estimate expression and identity features through a neural architecture composed of two identical CNN streams, in order to alleviate inter-subject variations introduced by personal attributes and to achieve better facial expression recognition performance. Liu et al.~\cite{Liu-CVPRW-2017} employed deep metric learning and jointly optimized a deep metric loss and the softmax loss. They obtained an identity-invariant model by using an identity-aware hard-negative mining and online positive mining scheme. Li et al.~\cite{Li-CVPR-2017} trained a CNN model using a modified back-propagation algorithm which creates a locality preserving loss aiming to pull the locally neighboring faces of the same class together. Li et al.~\cite{Li-ICPR-2018} proposed an end-to-end trainable Patch-Gated CNN that can automatically perceive occluded region of the face, making the recognition based on the visible regions. To find the visible regions of the face, their model decomposes an intermediate feature map into several patches according to the positions of related facial landmarks. Each patch is then reweighted by its importance, which is determined from the patch itself. Zeng et al.~\cite{Zeng-ECCV-2018} proposed a model that addresses the labeling inconsistencies across data sets. In their framework, images are tagged with multiple (pseudo) labels either provided by human annotators or predicted by learned models. Then, a facial expression recognition model is trained to fit the latent truth from the inconsistent pseudo-labels. Hua et al.~\cite{Hua-Access-2019} proposed a deep learning algorithm consisting of three sub-networks of different depths. Each sub-network is based on an independently-trained CNN. Different from Hua et al.~\cite{Hua-Access-2019}, we combine deep CNN features with handcrafted features and employ local learning.

Closer to our work are methods~\cite{Connie-MIWAI-2017,Kaya-IVC-2017} that combine deep and handcrafted features or that employ local learning~\cite{Ionescu-WREPL-2013} for facial expression recognition. While Ionescu et al.~\cite{Ionescu-WREPL-2013} used local learning to improve the performance of a handcrafted model, we show that local learning can also improve performance when deep features are used in combination with handcrafted features. Remarkably, our top accuracy is almost $8\%$ better than the accuracy reported in~\cite{Ionescu-WREPL-2013}. Works that combine deep and handcrafted features usually employ a single CNN model and various handcrafted features, e.g. Connie et al.~\cite{Connie-MIWAI-2017} employed SIFT and dense SIFT, while Kaya et al.~\cite{Kaya-IVC-2017} employed SIFT, HOG and Local Gabor Binary Patterns (LGBP). On the other hand, we employ a single type of handcrafted features and we include various CNN architectures in the combination. Another important difference from works~\cite{Connie-MIWAI-2017,Kaya-IVC-2017} that combine deep and handcrafted features is that we employ local learning in the training stage. With these key changes, the empirical results indicate that our approach achieves better performance than the approach of Connie et al.~\cite{Connie-MIWAI-2017}. We do not compare with Kaya et al.~\cite{Kaya-IVC-2017}, since their approach is designed to work on video. Hence, they do not report results on static image data sets. In future work, we aim to extend our approach for video, which will enable a direct comparison to Kaya et al.~\cite{Kaya-IVC-2017}. Nevertheless, our experiments on static images indicate that the proposed model combination achieves superior results than various state-of-the-art approaches~\cite{Barsoum-ICMI-2016,Connie-MIWAI-2017,Hua-Access-2019,Ionescu-WREPL-2013,Kim-JMUI-2016,Li-MTA-2017,Li-ICPR-2018,Mollahosseini-TAC-2017,Tang-WREPL-2013,Yu-ICMI-2015,Zeng-ECCV-2018}.


\section{Approach}
\label{sec_M}

\begin{figure*}[!t]
\begin{center}
\includegraphics[width=0.68\linewidth]{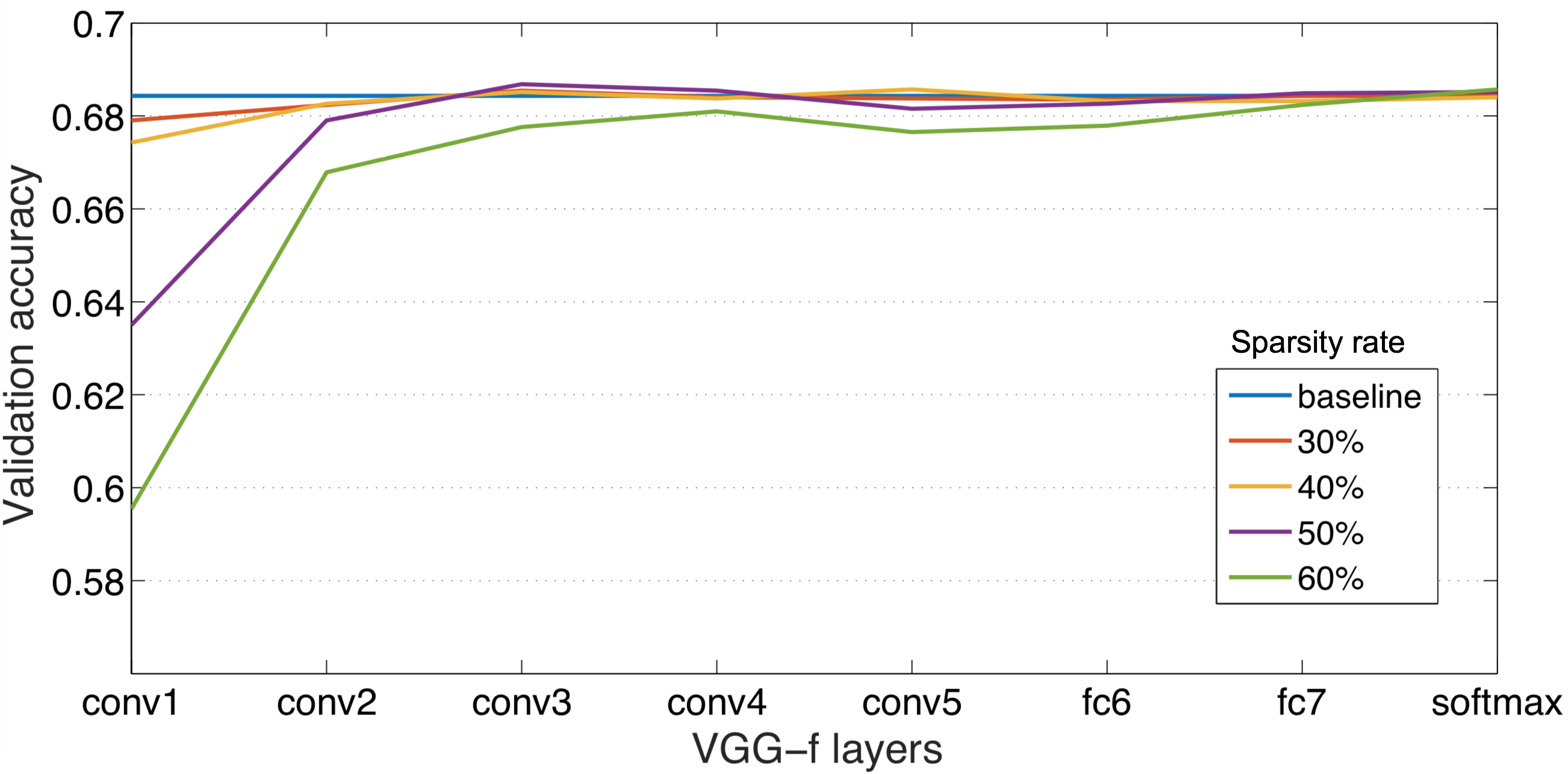}
\end{center}
\vspace{-0.2cm}
\caption{Validation accuracy rates of the VGG-f network, which is fine-tuned on FER 2013, after pruning the smaller weights on each individual layer using several sparsity rates between $30\%$ and $60\%$. The baseline represents the accuracy of the fine-tuned model without pruning, i.e. having a sparsity rate of $0\%$. The layers closer to the input are more sensitive to pruning. Best viewed in color.}
\label{fig_VGGf_sparse}
\vspace{-0.2cm}
\end{figure*}

\subsection{Deep Models}

We employ three CNN models in this work, namely VGG-face~\cite{Parkhi-BMVC-2015}, VGG-f~\cite{Chatfield-BMVC-14} and VGG-13~\cite{Barsoum-ICMI-2016}. Among these three models, only VGG-13 is trained from scratch. For the other two CNN models, we use pre-trained as well as fine-tuned versions. In order to train or fine-tune the models, we use stochastic gradient descent using mini-batches of $512$ images and the momentum rate set to $0.9$. All models are trained using data augmentation, which is based on including horizontally flipped images. To prevent overfitting, we employ Dense-Spare-Dense (DSD) training~\cite{Han-ICLR-2017} to train our CNN models. The training starts with a dense phase, in which the network is trained as usual. When switching to the sparse phase, the weights that have lower absolute values are replaced by zeros after every epoch. A sparsity threshold is used to determine the percentage of weights that are replaced by zeros. The DSD learning process, typically ends with a dense phase. It is important to note that DSD can be applied several times in order to achieve the desired performance.

\noindent
{\bf VGG-face.}
With $16$ layers, VGG-face~\cite{Parkhi-BMVC-2015} is the deepest network that we fine-tune. Since VGG-face is pre-trained on a closely related task (face recognition), we freeze the weights in the convolutional (conv) layers and we train only the fully-connected (fc) layers to adapt the network for our task (facial expression recognition). We replace the softmax layer of $1000$ units with a softmax layer of $7$ or $8$ units, depending on the data set, e.g. FER 2013~\cite{Goodfellow-ICONIP-2013} contains $7$ classes of emotion, while FER+~\cite{Barsoum-ICMI-2016} contains $8$ classes of emotion. We randomly initialize the weights in this layer, using a Gaussian distribution with zero mean and $0.1$ standard deviation. We add a dropout layer after the first fc layer, with the dropout rate set to $0.7$. We set the learning rate to $10^{-4}$ and we decrease it by a factor of $10$ when the validation error stagnates for more than $10$ epochs. We fine-tune VGG-face using DSD training~\cite{Han-ICLR-2017}. We train the network for $200$ epochs in a first dense phase. We then switch to a sparse phase and we carry on training for another $50$ epochs, with the sparsity rate set to $0.6$ for all fc layers. In the second dense phase, we train the network for $50$ epochs. We train the network for another $50$ epochs during a second sparse phase, without changing the sparsity rate. Finally, we train the network for another $50$ epochs during a third dense phase. In total, the network is trained for $400$ epochs.

\noindent
{\bf VGG-f.}
We also fine-tune the VGG-f~\cite{Chatfield-BMVC-14} network with $8$ layers, which is pre-trained on ILSVRC~\cite{Russakovsky2015}. Since VGG-f is pre-trained on a distantly related task (object class recognition), we fine-tune all of its layers. We set the learning rate to $10^{-4}$ and we decrease it by a factor of $10$ when the validation error stagnates for more than $10$ epochs. In the end, the learning rate drops to $10^{-5}$. After each fc layer, we add a dropout layer with the dropout rate set to $0.5$. We also add dropout layers after the last two conv layers, setting their dropout rates to $0.35$. In total, there are four dropout layers. As for VGG-face, we use the DSD training method to fine-tune the VGG-f model. However, we refrain from pruning the weights of the first two conv layers during the sparse phases, since pruning these layers has a higher negative impact on the validation accuracy of the network, as illustrated in Figure~\ref{fig_VGGf_sparse}. Based on the sensitivity analysis presented in Figure~\ref{fig_VGGf_sparse}, we choose, for each layer, the highest sparsity rate in the set $\{0.3,0.4,0.5,0.6\}$ that does not affect the validation accuracy by more than $0.5\%$. We train this network for a total of $600$ epochs using DSD. We start with a dense phase of $300$ epochs, then we alternate between sparse and dense phases, each phase lasting for $50$ epochs.

\begin{figure*}[!t]
\centering
\subfloat[A global linear classifier misclassifies the test samples depicted in red.]
{
	\includegraphics[width=0.35\linewidth]{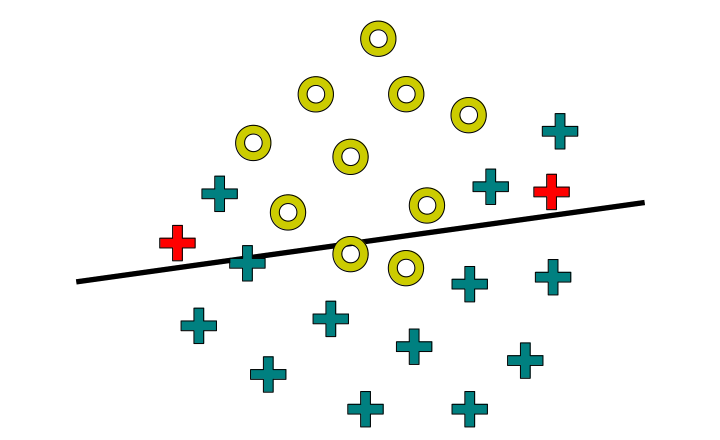}
    \label{Fig_LL_linear:subfig1} 
}
\hspace{0.06\linewidth}
\subfloat[Local learning models based on an underlying linear classifier are able correctly classify the test samples depicted in red. The grey area around each test sample represents the neighborhood of the respective test sample.]
{
	\includegraphics[width=0.35\linewidth]{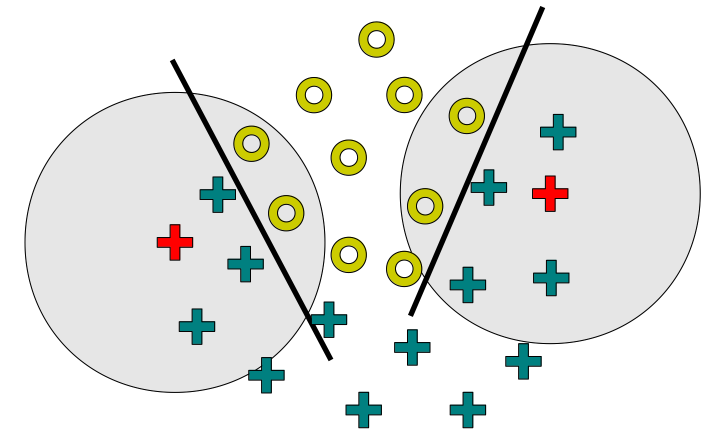}
    \label{Fig_LL_nonlinear:subfig1} 
}
\caption{Two classification models are used to solve the same binary classification problem. The two test samples depicted in red are misclassified by a global linear classifier (left-hand side). The local learning\index{local learning} framework produces a non-linear decision boundary that fixes this problem (right-hand side). Best viewed in color.}
\label{Fig_Local_Nonlinear}
\vspace{-0.2cm}
\end{figure*}

\begin{figure*}[!t]
\begin{center}
\includegraphics[width=0.92\linewidth]{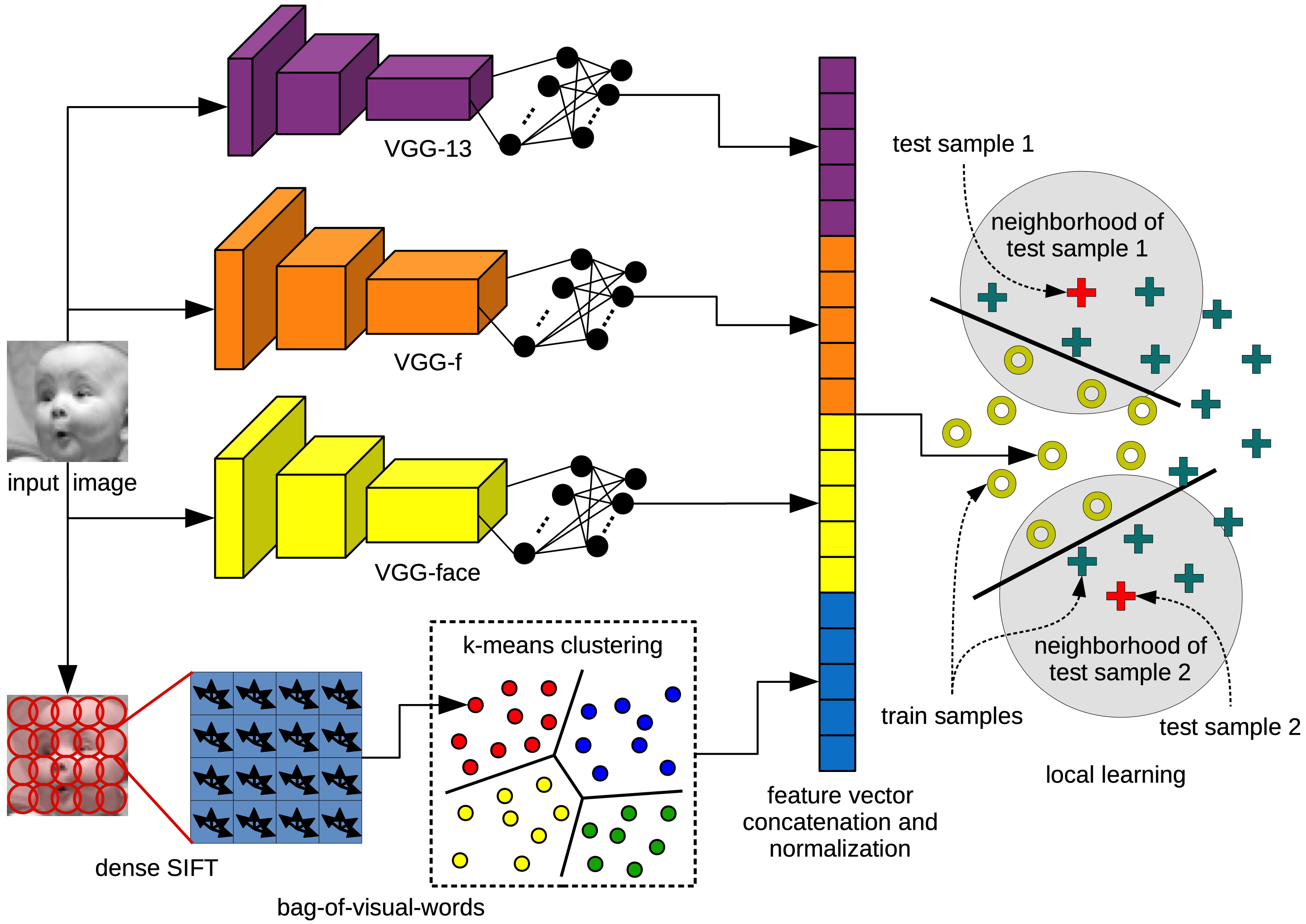}
\end{center}
\vspace{-0.2cm}
\caption{Our processing pipeline based on automatic features learned by convolutional neural networks (VGG-13, VGG-f and VGG-face) and handcrafted features computed by the bag-of-visual-words model. After feature vector concatenation and $L_2$-normalization, we employ a local learning model. Best viewed in color.}
\label{fig_pipeline}
\vspace{-0.2cm}
\end{figure*}

\noindent
{\bf VGG-13.}
The VGG-13 architecture was specifically designed by Barsoum et al.~\cite{Barsoum-ICMI-2016} for the FER+ data set. Since the images in FER 2013 are of the same size, we consider that VGG-13 is an excellent choice for FER 2013 as well. Although the model is not particularly adapted for the AffectNet data set~\cite{Mollahosseini-TAC-2017}, which contains larger images, we keep the same architecture for all data sets. The weights are randomly initialized, by drawing them from a Gaussian distribution with zero mean and $0.01$ standard deviation. We use the same dropout rates as in the original paper~\cite{Barsoum-ICMI-2016}. We set the initial learning rate to $10^{-2.5}$ and we decrease it by a factor of $10^{0.5}$ whenever the validation error stops decreasing. The last learning rate that we use is $10^{-4}$. We train VGG-13 for $100$ epochs using dense training. We then switch to a sparse phase that lasts for $50$ epochs, with the sparsity rate set to $0.1$. The training ends with a second dense phase that lasts for $50$ epochs. In total, the VGG-13 network is trained for $200$ epochs. It is important to mention that, different from Barsoum et al.~\cite{Barsoum-ICMI-2016}, we use the softmax loss instead of probabilistic label drawing to train VGG-13. Hence, the results of the individual VGG-13 model are slightly different than those reported in~\cite{Barsoum-ICMI-2016}.

\subsection{Handcrafted Model}

The BOVW model proposed for facial expression recognition is divided in two pipelines, one for training and one for testing. In the training pipeline, we build the feature representation by extracting dense SIFT descriptors~\cite{bosch-phow-2007,Lowe-SIFT-2004} from all training images, and by later quantizing the extracted descriptors into visual words using k-means clustering~\cite{Leung-2001}. The visual words are then stored in a randomized forest of k-d trees~\cite{Philbin-2007} to reduce search cost. After building the vocabulary of visual words, the training and testing pipelines become equivalent. For each image in the training or testing sets, we record the presence or absence of each visual word in a binary feature vector. The standard BOVW model described so far ignores spatial relationships among visual words, but we can achieve better performance by including spatial information. Perhaps the most popular and straightforward approach to include spatial information is the spatial pyramid~\cite{Lazebnik-BBF-2006}. Our spatial pyramid representation is obtained by dividing the image into increasingly fine sub-regions (bins) and by computing the binary feature vector corresponding to each bin. The final representation is a concatenation of all binary feature vectors. It is reasonable to think that dividing an image representing a face into bins is a good choice, since most features, such as the contraction of the muscles at the corner of the eyes, are only visible in a certain region of the face. 

\subsection{Model Fusion and Learning}

\noindent
{\bf Model fusion.} We combine the deep and handcrafted models before the learning stage, by concatenating the corresponding features. To extract deep features from the pre-trained or fine-tuned CNN models, we remove the softmax classification layer and we consider the activation map of last remaining fc layer as the deep feature vector corresponding to the image provided as input to the network. The deep feature vectors are normalized using the $L_2$-norm. The bag-of-visual-words representation is the only kind of handcrafted features that we employ. The BOVW feature vectors are also normalized using the $L_2$-norm. Our full processing pipeline is illustrated in Figure~\ref{fig_pipeline}.

\noindent
{\bf Global learning.}
We employ the linear Support Vector Machines (SVM)~\cite{cortes-vapnik-ml-1995} to learn a discriminative model based on all training examples. SVM is a binary classifier that tries to find the vector of weights and the bias term that define the hyperplane which maximally separates the feature vectors of the training examples belonging to the two classes. 
To extend the linear SVM classifier to our multi-class facial expression recognition problem, we employ the one-versus-all scheme.

\noindent
{\bf Local learning.}
Local learning methods attempt to locally adjust the performance of the training system to the properties of the training set, in each area of the input space. A local learning algorithm essentially works by $(i)$ selecting a few training samples located in the vicinity of a given test sample, then by $(ii)$ training a classifier with only these few examples and finally, by $(iii)$ applying the classifier to predict the class label of the test sample.

It is interesting to note that the k-nearest neighbors (k-NN) model can be included in the family of local learning algorithms. Actually, the k-NN model is the simplest formulation of local learning, since the discriminant function is constant (there is no learning involved). What is even more interesting, however, is that almost any other classifier can be employed in the local learning paradigm. In our case, we employ the linear SVM classifier for the local classification problem. 

It is important to mention that besides the classifier, a similarity or distance measure is also required to determine the neighbors located in the vicinity of a test sample. In our case, we use the cosine similarity. 

An interesting remark is that a linear classifier, such as SVM, put in the local learning framework, becomes non-linear, as shown in Figure~\ref{Fig_Local_Nonlinear}. In the standard approach, a single linear classifier trained at the global level (on the entire train set) produces a linear discriminative function. On the other hand, the discriminative function for a set of test samples is no longer linear in the local learning framework, since each prediction is given by a different linear classifier which is specifically trained for a single test sample. Moreover, the discriminative function cannot be determined without having the test samples beforehand, yet the local learning paradigm is able to rectify some limitations of linear classifiers, as illustrated in Figure~\ref{Fig_Local_Nonlinear}. Local learning\index{local learning} has a few advantages over standard learning methods. First, it divides a hard classification problem into multiple simple sub-problems. Second, it reduces the variety of samples in the training set, by selecting the samples that are most similar to the test sample. 

\section{Experiments}
\label{sec_E}

\subsection{Data Sets}

We conduct experiments on the FER 2013~\cite{Goodfellow-ICONIP-2013}, the FER+~\cite{Barsoum-ICMI-2016} and the AffectNet~\cite{Mollahosseini-TAC-2017} data sets. 

\noindent
{\bf FER 2013.}
The FER 2013 data set contains $28709$ training images, $3589$ validation (public test) images and another $3589$ (private) test images. All images are of $48 \times 48$ pixels in size. The images belong to $7$ classes of emotion: \textit{anger}, \textit{disgust}, \textit{fear}, \textit{happiness}, \textit{neutral}, \textit{sadness}, \textit{surprise}. 

\noindent
{\bf FER+.}
The FER+ data set is a curated version of FER 2013 in which some of the original images are relabeled, while other images, e.g. not containing faces, are completely removed. Interestingly, Barsoum et al.~\cite{Barsoum-ICMI-2016} add \textit{contempt} as the eighth class of emotion. The FER+ data set contains $25045$ training images, $3191$ validation images and another $3137$ test images.

\noindent
{\bf AffectNet.}
The AffectNet~\cite{Mollahosseini-TAC-2017} data set contains $287651$ training images and $4000$ validation images, which are manually annotated. Since the test set is not publicly available, researchers~\cite{Mollahosseini-TAC-2017,Zeng-ECCV-2018} evaluate their approaches on the validation set containing $500$ images for each of the following 8 emotion classes: \textit{anger}, \textit{contempt}, \textit{disgust}, \textit{fear}, \textit{happiness}, \textit{neutral}, \textit{sadness}, \textit{surprise}. As the facial expression recognition task typically includes only 7 emotion classes (\textit{contempt} is excluded), some works~\cite{Hua-Access-2019,Li-ICPR-2018} report results on $3500$ validation images from AffectNet, by removing the $500$ images labeled with the \textit{contempt} emotion. We evaluate our approach in both 8-way and 7-way classification settings, in order to provide a comprehensive comparison with related works~\cite{Hua-Access-2019,Li-ICPR-2018,Mollahosseini-TAC-2017,Zeng-ECCV-2018}.

\begin{table*}[!t]
\setlength\tabcolsep{4.5pt}
\small{
\caption{Results on the FER 2013~\cite{Goodfellow-ICONIP-2013}, the FER+~\cite{Barsoum-ICMI-2016} and the AffectNet~\cite{Mollahosseini-TAC-2017} data sets. Our combination based on pre-trained, fine-tuned and handcrafted models, with and without data augmentation (aug.), are compared with several state-of-the-art approaches~\cite{Barsoum-ICMI-2016,Connie-MIWAI-2017,Hua-Access-2019,Ionescu-WREPL-2013,Kim-JMUI-2016,Li-MTA-2017,Li-ICPR-2018,Mollahosseini-TAC-2017,Tang-WREPL-2013,Yu-ICMI-2015,Zeng-ECCV-2018}
, which are listed in temporal order. The best result on each data set is highlighted in bold.
}\label{tab_results}
\vspace{-0.2cm}
\begin{center}
\begin{tabular}{|l||c|c||c|c||c|c||c|c|}
\hline
Model                                           							& FER		& FER	    				& FER+	     & FER+ 	&AffectNet 	& AffectNet		&AffectNet 	& AffectNet \\
                                           									& 	        		& (aug.)	    			&				& (aug.) 	& 8-way 		& 8-way (aug.)	& 7-way		& 7-way (aug.)\\
\hline
\hline
Ionescu et al.~\cite{Ionescu-WREPL-2013}        & $67.48\%$     & -                 & -             & - 					& -             & -				& -             & -\\
\hline
Tang~\cite{Tang-WREPL-2013}                     		& -             & $71.16\%$         	& -             & -					& -             & -				& -             & -\\
\hline
Yu et al.~\cite{Yu-ICMI-2015}                    			& -             & $72.00\%$         & -             & -					& -             & -				& -             & -\\
\hline
Kim et al.~\cite{Kim-JMUI-2016}                   		& -             & $72.72\%$         & -             & -					& -             & -				& -             & -\\
\hline
Barsoum et al.~\cite{Barsoum-ICMI-2016}          & -             & -                 		& -             & $84.99\%$	& -             & -				& -             & -\\
\hline
Li et al.~\cite{Li-MTA-2017}                      			& -             & $70.66\%$         & -             & -					& -            & -				& -             & -\\
\hline
Connie et al.~\cite{Connie-MIWAI-2017}            & -             & $73.40\%$         & -             & -					& -            & -				& -             & -\\
\hline
Mollahosseini et al.~\cite{Mollahosseini-TAC-2017} & -     & -        					& -             & -					& -            & $58.00\%$	& -             & -\\
\hline
Li et al.~\cite{Li-ICPR-2018}								& -     		& -        					& -             & -					& - 			& - 				& $55.33\%$ & - \\
\hline
Zeng et al.~\cite{Zeng-ECCV-2018}					& -     		& -        					& -             & -					& -			& $57.31\%$	 & -             & -\\
\hline
Hua et al.~\cite{Hua-Access-2019}					& -			& $71.91\%$ 			& -             & -					& - 			& - 				& - 			& $62.11\%$\\
\hline
\hline
CNNs and BOVW  + global SVM	& $73.34\%$     & $73.25\%$         	& $86.68\%$     & $86.96\%$ 				& $59.20\%$    & $59.30\%$ 	& $63.20\%$      & $62.91\%$\\
\hline
CNNs and BOVW + local SVM		& $74.92\%$     & $\mathbf{75.42\%}$& $\mathbf{87.76\%}$& $87.25\%$ 	& $59.45\%$    & $\mathbf{59.58\%}$ & $62.94\%$  & $\mathbf{63.31\%}$\\
\hline
\end{tabular}
\end{center}
}
\end{table*}

\begin{figure*}[!th]
\begin{center}
\includegraphics[width=0.68\linewidth]{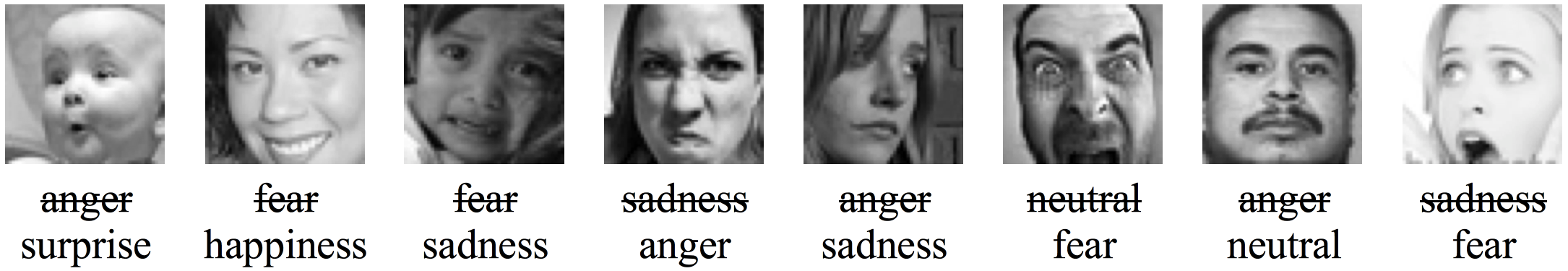}
\end{center}
\vspace{-0.2cm}
\caption{FER 2013 test images that are incorrectly predicted by the global SVM based on our combination of deep and handcrafted features, but are correctly predicted by the local SVM based on the same feature combination.}
\label{fig_fer_examples}
\vspace{-0.2cm}
\end{figure*}

\subsection{Implementation Details}

The input images are scaled to $224 \times 224$ pixels for VGG-face and VGG-f, and to $64 \times 64$ pixels for VGG-13. We use the MatConvNet~\cite{matconvnet} library to train the CNN models. To implement the BOVW model, we use functions from the VLFeat~\cite{vedaldi-vlfeat-2008} library. To generate the spatial pyramid representation for the BOVW model, we divide the images into $1\times1$, $2\times2$, $3\times3$ and $4\times4$ bins. At each level of the pyramid, we use vocabularies of $17000$, $14000$, $11000$ and $8000$ words, respectively. In the training phase, we employ the SVM implementation from LibSVM~\cite{LibSVM-2011}. We set the regularization parameter of SVM to $C=1$ for individual models and to $C=100$ for combined models. We employ the linear kernel, which does not require any additional parameters. In the local learning approach, we employ the cosine similarity to choose the nearest neighbors. We select $200$ neighbors for training the local SVM. All parameters are tuned on the validation sets from FER 2013 and FER+. We transfer the parameter values to AffectNet, without further tuning.

To train our deep or handcrafted models on AffectNet, we adopt the down-sampling setting proposed in~\cite{Mollahosseini-TAC-2017}, which solves, to some extent, the imbalanced nature of the facial expression recognition task. As Mollahosseini et al.~\cite{Mollahosseini-TAC-2017}, we select at most $15000$ samples from each class. This leaves us with a training set of $88021$ images.

We use the same model combination on all data sets. The proposed combination includes the BOVW representation and the deep features extracted with pre-trained VGG-face, fine-tuned VGG-face, fine-tuned VGG-f and VGG-13. The combination is obtained by concatenating the corresponding features.

\subsection{Results}

Table~\ref{tab_results} includes the results of our combined models, one based on global SVM and another based on local SVM, on three data sets: FER 2013, FER+ and AffectNet. We report results with and without data augmentation. Our models are compared with several state-of-the-art approaches~\cite{Barsoum-ICMI-2016,Connie-MIWAI-2017,Hua-Access-2019,Ionescu-WREPL-2013,Kim-JMUI-2016,Li-MTA-2017,Li-ICPR-2018,Mollahosseini-TAC-2017,Tang-WREPL-2013,Yu-ICMI-2015,Zeng-ECCV-2018}.

\noindent
{\bf Results on FER 2013.}
With our combination of features, the local SVM classifier achieves an accuracy rate of $74.92\%$ when the FER 2013 training set is not augmented with flipped images, and an accuracy rate of $75.42\%$ when the training set is augmented with flipped images. In the latter case, the difference from the global SVM is $2.17\%$. We consider that the trade-off between accuracy and speed is acceptable, given that the local SVM finds the nearest neighbors and predicts the test labels in $40.28$ seconds for all $3589$ test images, while the global SVM predicts the labels in $23.93$ seconds. The reported running times are measured on a computer with Intel Xeon $2.20$ GHz Processor and $256$ GB of RAM, using a single thread.

Figure~\ref{fig_fer_examples} provides a handful of test images that are incorrectly labeled by the global SVM, but correctly labeled by the local SVM. We also tried to determine if applying SVM locally (on the selected nearest neighbors) is indeed helpful in comparison with a k-NN model. The k-NN model yields an accuracy of $70.33\%$ with the same number of neighbors ($200$). We thus conclude that the local SVM approach provides a considerable improvement over the k-NN model. Moreover, with a top accuracy of $75.42\%$, we surpass the accuracy of the state-of-the-art model~\cite{Connie-MIWAI-2017} by $2.02\%$. 

\noindent
{\bf Results on FER+.} In both settings (with and without data augmentation), the local SVM approach yields better accuracy rates than the global SVM approach on FER+, but the differences are not as high as for FER 2013. Without data augmentation, our combination of deep and handcrafted features attains an accuracy of $86.68\%$ when the global SVM is employed in the training phase, and an accuracy of $87.76\%$ when the local SVM is used instead of the global SVM. The local SVM classifier attains an accuracy improvement of $1.08\%$ over the global SVM. Data augmentation does not seem to help us gain any performance improvements on FER+, but it is important to note that the local SVM still attains better performance than the global SVM. In the end, we surpass the state-of-the-art method~\cite{Barsoum-ICMI-2016} on the FER+ data set by $2.77\%$, reaching the best accuracy of $87.76\%$ using the local SVM without data augmentation.

\begin{table*}[!t]
\setlength\tabcolsep{4.5pt}
\small{
\caption{Ablation results on the FER 2013~\cite{Goodfellow-ICONIP-2013}, the FER+~\cite{Barsoum-ICMI-2016} and the AffectNet~\cite{Mollahosseini-TAC-2017} data sets. Our combination of deep and handcrafted models is compared with each individual component of the combination. Results are reported with and without data augmentation (aug.). The best result on each data set is highlighted in bold.
}\label{tab_ablation}
\vspace{-0.2cm}
\begin{center}
\begin{tabular}{|l||c|c||c|c||c|c||c|c|}
\hline
Model                                           		& FER				& FER	    			& FER+	     		& FER+ 			&AffectNet 		& AffectNet		&AffectNet 		& AffectNet \\
                                           				& 	        				& (aug.)	    		&						& (aug.) 			& 8-way 			& 8-way (aug.)	& 7-way			& 7-way (aug.)\\
\hline
\hline
BOVW 											& $65.70\%$	& $66.23\%$	& $79.60\%$	& $80.65\%$	& $47.53\%$	& $48.30\%$	& $51.51\%$		& $52.29\%$\\
\hline
pre-trained VGG-face 					& $65.65\%$	& $65.78\%$	& $81.54\%$		& $81.73\%$		& $49.28\%$	& $50.08\%$	& $54.14\%$		& $54.94\%$\\
\hline
fine-tuned VGG-face                       & $71.50\%$	    & $72.11\%$		& $84.35\%$	& $84.79\%$	& $58.77\%$		& $58.93\%$	& $62.54\%$	& $62.66\%$\\
\hline
fine-tuned VGG-f                         	& $69.38\%$	& $70.30\%$	& $85.72\%$		& $86.01\%$		& $55.85\%$	& $56.03\%$	& $60.40\%$	& $60.51\%$\\
\hline
VGG-13                              				& $66.31\%$     & $66.51\%$		& $84.38\%$	& $84.41\%$		& $40.50\%$	& $41.75\%$		& $44.60\%$	& $44.57\%$\\
\hline
\hline
CNNs and BOVW  + global SVM	& $73.34\%$     & $73.25\%$	& $86.68\%$	& $86.96\%$ 	& $59.20\%$    & $59.30\%$ 	& $63.20\%$     & $62.91\%$\\
\hline
CNNs and BOVW + local SVM		& $74.92\%$   & $\mathbf{75.42\%}$& $\mathbf{87.76\%}$		& $87.25\%$ 	& $59.45\%$    & $\mathbf{59.58\%}$ & $62.94\%$  & $\mathbf{63.31\%}$\\
\hline
\end{tabular}
\end{center}
}
\end{table*}

\noindent
{\bf Results on AffectNet 8-way.}
First, we note that Mollahosseini et al.~\cite{Mollahosseini-TAC-2017} attained better results using weighted-loss ($58.00\%$) instead of down-sampling ($50.00\%$) the AffectNet training set. Although we use down-sampling to train our models, we compare our results with the better (weighted-loss) version of Mollahosseini et al.~\cite{Mollahosseini-TAC-2017}. We are able to surpass their approach by $1.58\%$, reaching an accuracy of $59.58\%$ on AffectNet with our local SVM based on the combination of deep and handcrafted features. We note that the local SVM attains superior results compared to the global SVM, in both settings, i.e. with and without data augmentation.

\noindent
{\bf Results on AffectNet 7-way.}
Some researchers~\cite{Hua-Access-2019,Li-ICPR-2018} reported results on AffectNet, by excluding the $500$ images labeled as \textit{contempt}. We include a comparison with these works in Table~\ref{tab_results}. When we do not use data augmentation, we notice that the global SVM outperforms the local SVM. However, the local SVM approach is better than the global SVM, when data augmentation is included. Our best result on the AffectNet 7-way classification task ($63.31\%$) is obtained by the local SVM that includes data augmentation. Our accuracy is $1.20\%$ higher than the state-of-the-art accuracy reported in~\cite{Hua-Access-2019}.

\noindent
{\bf Results overview.}
The empirical results presented in Table~\ref{tab_results} show that the local SVM model based on our combination of deep and handcrafted features achieves superior performance compared to several recent and related works~\cite{Barsoum-ICMI-2016,Connie-MIWAI-2017,Hua-Access-2019,Ionescu-WREPL-2013,Kim-JMUI-2016,Li-MTA-2017,Li-ICPR-2018,Mollahosseini-TAC-2017,Tang-WREPL-2013,Yu-ICMI-2015,Zeng-ECCV-2018}. We also note that the local SVM generally attains better performance than the global SVM, in all but one case, proving that the idea of using local learning is indeed useful. Overall, the results demonstrate that our method based on local SVM and a combination of deep and handcrafted features, achieves top performance on all three data sets: FER 2013, FER+ and AffectNet.

\subsection{Ablation Results}

Table~\ref{tab_ablation} includes the results of our combined models, one based on global SVM and another based on local SVM, in comparison with each and every individual component, on three data sets: FER 2013, FER+ and AffectNet.

\noindent
{\bf BOVW.}
The accuracy rates of our BOVW model, which is based on global SVM, are generally lower than the accuracy rates of the deep CNN models. It seems that the BOVW model is able to surpass VGG-13 on AffectNet. However, we believe that this happens only because the VGG-13 architecture is not specifically adapted to the larger AffectNet images.

\noindent
{\bf VGG-face.}
Although the pre-trained VGG-face~\cite{Parkhi-BMVC-2015} is trained on a rather complementary task, face recognition, it achieves fairly good accuracy rates, e.g. $81.73\%$ on FER+. Fine-tuning the VGG-face model using data augmentation improves its accuracy rates on all data sets, usually by a very large margin (up to $9\%$ over the pre-trained VGG-face). 

\noindent
{\bf VGG-f.}
Since VGG-f~\cite{Chatfield-BMVC-14} is pre-trained on a distantly-related task, object class recognition, we do not consider it as a viable model to be included in our combination. However, the fine-tuned VGG-f model reaches respectable accuracy rates (see Table~\ref{tab_ablation}), even surpassing the fine-tuned VGG-face model on FER+. We also note that our VGG-f model trained on AffectNet using down-sampling attains an accuracy of $56.03\%$, surpassing the AlexNet model of Mollahosseini et al.~\cite{Mollahosseini-TAC-2017} trained using down-sampling, which attains an accuracy of $50.00\%$. Although the two networks, VGG-f and AlexNet, have fairly similar architectures, we believe that the significant performance difference between these models is due to the DSD training procedure, which we applied for training all our CNN models, including VGG-f.

\noindent
{\bf VGG-13.}
The VGG-13~\cite{Barsoum-ICMI-2016} model, which is trained from scratch, achieves an accuracy of $66.51\%$ on FER 2013 and an accuracy of $84.41\%$ on FER+. Since the input of the VGG-13 architecture is $64\times64$ pixels in size, it seems to be better suited to the FER 2013 or the FER+ data sets, both containing images of $48\times48$ pixels, compared to the VGG-face or the VGG-f architectures, which take as input images of $224\times224$ pixels. However, its lower performance compared to VGG-face or VGG-f can be explained by the fact that the other CNN models have a better starting point, since they are pre-trained on related computer vision tasks. It is interesting to note that our own implementation of the VGG-13 architecture of Barsoum et al.~\cite{Barsoum-ICMI-2016} attains an accuracy of $84.41\%$ on FER+, which is $0.58\%$ less than the accuracy reported in~\cite{Barsoum-ICMI-2016}. We believe that this difference is a consequence of using the standard softmax loss function instead of probabilistic label drawing. 

\noindent
{\bf Ablation study overview.}
With an accuracy of $72.11\%$, the best individual model on FER 2013 is the fine-tuned VGG-face. We note that all models obtain much better results on FER+ than on FER 2013, indicating that the FER+ curation process conducted by Barsoum et al.~\cite{Barsoum-ICMI-2016} was indeed helpful. Although the fine-tuned VGG-face obtains better results than the other fine-tuned networks on FER 2013, it seems that the shallower VGG-f reaches the best performance ($86.01\%$) among individual models, when it is fine-tuned on FER+. As for FER 2013, the best individual model on AffectNet is the fine-tuned VGG-face. In the AffectNet 8-way classification task, it achieves an accuracy of $58.93\%$, and in the AffectNet 7-way classification task, it achieves an accuracy of $62.66\%$. Overall, the ablation results presented in Table~\ref{tab_ablation} show that our model combination provides better results than each individual counterpart. We thus conclude that our combination of deep and handcrafted features is indeed necessary to improve performance over each component of the combination.

\section{Conclusion}
\label{sec_C}

In this paper, we have presented a state-of-the-art approach for facial expression recognition, which is based on combining deep and handcrafted features and on applying local learning in the training phase. With a top accuracy of $75.42\%$ on FER 2013, a top accuracy of $87.76\%$ on FER+, a top accuracy of $59.58\%$ on AffectNet 8-way classification and a top accuracy of $63.31\%$ on AffectNet 7-way classification, our approach is able to surpass the best methods on these data sets~\cite{Barsoum-ICMI-2016,Connie-MIWAI-2017,Hua-Access-2019,Mollahosseini-TAC-2017}. 

In future work, we aim to evaluate our approach on additional data sets and adapt our method for video. We also consider training our approach to distinguish between voluntary (deceptive) and involuntary (natural) facial expressions.

\section*{Acknowledgments}
This research is partially supported by Novustech Services through Project 115788 funded under the Competitiveness Operational Programme POC-46-2-2.

{\small
\bibliographystyle{ieee}
\bibliography{references}
}

\end{document}